%% file: iclr2023_conference.tex
\documentclass{article} 
\usepackage{iclr2023_conference,times}

\input{math_commands.tex}

\pdfoutput=1

\usepackage{hyperref}
\usepackage{url}
\usepackage{tikz}
\usepackage{xcolor}
\usepackage{tcolorbox}
\usepackage{color,xcolor}

\usepackage{amssymb}
\usepackage{amsopn}
\usepackage{graphicx}
\usepackage{multirow}
\usepackage[english]{babel}
\usepackage{bm}
\usepackage{array}
\usepackage{setspace}
\usepackage{todonotes}

\usepackage{xspace}
\usepackage{amsfonts}
\usepackage{array,multirow}
\usepackage{pgfplots}
\usepackage{tikz}
\usepackage{verbatim}
\usepackage{subfig}
\usepackage{arydshln}
\usepackage{booktabs}
\usepackage{fancyhdr}

\title{ChuXin: 1.6B Technical Report}

\author{{\bf Xiaomin Zhuang$^{1}$\thanks{Equal contribution, correspondence to 
\{xiaominzhuang\}@outlook.com}  , \ Yufan Jiang$^{1*}$, \ Qiaozhi He$^{1}$, \ Zhihua Wu$^{1}$,} \\
}

\iclrfinalcopy 
\begin{document}

\maketitle

\begin{abstract}
In this report, we present ChuXin, an entirely open-source language model with a size of 1.6 billion parameters. Unlike the majority of works that only open-sourced the model weights and architecture, we have made everything needed to train a model available, including the training data, the training process, and the evaluation code. Our goal is to empower and strengthen the open research community, fostering transparency and enabling a new wave of innovation in the field of language modeling. 
Furthermore, we extend the context length to 1M tokens through lightweight continual pretraining and demonstrate strong needle-in-a-haystack retrieval performance. The weights for both models are available at Hugging Face to download and use\footnote{https://huggingface.co/chuxin-llm/Chuxin-1.6B-Base} \footnote{https://huggingface.co/chuxin-llm/Chuxin-1.6B-1M}.

\end{abstract}

\section{Introduction}
Large language models (LLMs)\citep{brown2020language,openai2023gpt4,touvron2023llama,chowdhery2022palm,biderman2023pythia,smith2022using,jiang2023mistral,yang2023baichuan,young2024yi} have revolutionized the field of natural language generation for their abilities to generate satisfactory text across various application domains.
Countless models have been open-sourced on AI communities like HuggingFace to facilitate their use by researchers \citep{bai2023qwen,singer2024h2o,zhang2024tinyllama}.
These models can broadly be divided into two categories: 1) Open source model weights and data sources, which constitute the vast majority. 2) Open source of all information about the model, including data used for training, data sampling ratios, intermediate checkpoints, and training logs, as well as evaluation methods, such as Tiny-Llama \citep{zhang2024tinyllama},OLMo \citep{groeneveld2024olmo} and StableLM 1.6B \citep{bellagente2024stable}.
Despite the performance of models released in the community continuing to make breakthroughs, we believe that full access to open language models for the research community is vital for thoroughly investigating these models' capabilities and limitations, as well as for understanding their inherent biases and potential risks.

In this report, we introduce ChuXin 1.6B, a model that is nearly fully open-sourced. ChuXin is trained entirely on open-source data, amounting to 2.3T tokens, which is a combination of web documents, encyclopedias and public knowledge databases in both English and Chinese. 
The details of the model's training procedure and the intermediate model checkpoints have been made publicly available.
ChuXin achieves comparable performance among open-source models of the same scale. 
This project has also been inspired by prior open-source efforts like OLMo \citep{groeneveld2024olmo}, Tiny-Llama \citep{zhang2024tinyllama} and StableLM 1.6B \citep{bellagente2024stable}.
Thus, we firmly believe that fostering a large and diverse ecosystem of open language models is the most promising path for advancing our scientific understanding of these models and driving technological innovations to enhance their practical utility.
Moreover, we have expanded the context length capability of ChuXin. To achieve an input length of 1M, we continued the pre-training of ChuXin on datasets sampled from longer texts.
we continue pretrain the model on length-upsampled data, following the method that used in the concurrent research by \citet{fu2024data}.
Ultimately, ChuXin also achieved satisfactory results on long-text test sets.
This is just the first step in our open-source journey. In the future, we plan to release larger models as well as instruction-tuned versions, while also incorporating more modalities. We will also document the various issues encountered during the training process and their solutions, providing the open-source language model community with more helpful and referential information.

This report is organized as follows: Section 2 details the process of pre-training ChuXin 1.6B. In Section 3, we showcase the model's evaluations based on established downstream benchmarks. Section 4 is dedicated to describing the process of long context extension. Finally, Section 5 concludes and summarizes this work.

\section{Pretraining}
 
\subsection{ Model and Architecture}
We select LLaMA2 \citep{touvron2023llama} as our backbone and adjust its architecture for a total of around 1.6B parameters. The model architectures are shown in Table \ref{tab_para}. In the following, we elaborate more details about the architecture of ChuXin 1.6B.
\begin{table*}[!ht]
\centering
\small
\begin{spacing}{1.2}
\setlength{\tabcolsep}{1.3mm}{
\begin{tabular}{lc}
\toprule
{\bf Pre-training Hyperparameters} & {\bf Value} \\
\midrule
Hidden size & 2048  \\
Intermediate size & 5632 \\
Layers & 24  \\
Heads & 32  \\
Sequence length & 4096 \\
Vocab size & 102400 \\
Attention variant  & full \\
Attention variant  & full \\
Biases & None \\
\bottomrule 
\end{tabular}}
\end{spacing}
\caption{
\label{tab_para}
{\bf ChuXin 1.6B model architecture.}
}
\end{table*}
\paragraph{Rotary positional embeddings (RoPE).}
To capture the relationships between sequence elements at distinct positions, we incorporate the Rotary Positional Embedding (RoPE) method, initially presented by \citet{su2024roformer}. 

\paragraph{RMSNorm.} We implement pre-normalization by normalizing the input before every sub-layer within the transformer for a more stable training process. Moreover, we utilize RMSNorm \citep{zhang2019root} for our normalization approach, enhancing the efficiency of the training process.

\paragraph{Attention Mask.} Inspired by stableLM \citep{bellagente2024stable}, we incorporated a block-diagonal attention mask design, which resets attention masks at EOS (End of Sequence) tokens across all packed sequences. This approach prevents cross-attention between disparate documents within the model's cool down phase, thereby further improving its performance.

\paragraph{Tokenizer.} We employed the DeepSeek LLM tokenizer \citep{bi2024deepseek}, which utilizes Byte-Level Byte-Pair Encoding (BBPE) from tokenizers library (Huggingface Team, 2019) for data tokenization. The vocabulary size is 102,400. The tokenizer was trained on a multilingual corpus of approximately 24 GB. Furthermore, this tokenizer is designed to split numbers into individual digits, enhancing the encoding of numeric data.

\paragraph{Further details.} We use SwiGLU \citep{shazeer2020glu} as our activation function and we do not use bias within linear layers nor tie word embeddings.

\subsection{Pretraining Data}
\begin{table*}[!ht]
\centering
\small
\begin{spacing}{1.1}
\setlength{\tabcolsep}{1.3mm}{
\begin{tabular}{lccc}
\toprule
{\bf Source} & {\bf Weight} & {\bf Num Tokens} & {\bf Category} \\
\midrule
RefinedWeb \citep{penedo2023refinedweb} & 0.29 & 493B & web \\
SlimPajama \citep{cerebras2023slimpajama} & 0.21 & 617B & web \\
Starcoderdata \citep{li2023starcoder} & 0.13 & 270B & code \\
Math & 0.03 & 15B & - \\
\quad OpenWebMath & & & math \\
\quad OpenMathInstruct  & & & math \\
\quad AMPS & & & math \\
\quad MathPile & & & math \\
En-others & 0.2 & 310B & - \\
\quad Arxiv &  &  & academic \\
\quad PubMed &  &  & academic \\
\quad S2ORC &  &  & academic \\
\quad PhilPapers &  &  & academic \\
\quad EuroParl &  & & law \\
\quad FreeLaw &  &  & law \\
\quad PileOfLaw &  &  & law \\
\quad RedPajama Wiki &  &  & Wiki \\
\quad Wiki(Dolma) &  &  & Wiki \\
\quad StackExchange &  &  & Social \\
\quad HackerNews &  &  & Social \\
\quad Reddit(Dolma) &  &   & Social \\
\quad Books(Dolma) &  &   & Books \\
\quad BookCorpusOpen &  &  & Books \\
\quad FanFics &  &  & Books \\
\quad PeS2o(Dolma) &  & & STEM papers \\
Cn-data & 0.11 & 580B & web,academic,books,encyclopedias \\
Instruct & 0.03 & 31B & instruction \\
\midrule
Total & 1.0 & 2.3T &  \\
\bottomrule 
\end{tabular}}
\end{spacing}
\caption{
\label{tab_data}
{\bf The complete training set with sampling weights.}}
\end{table*}
In order to make it easier for others to replicate our pre-trained model, all the pre-training datasets we used are sourced from HuggingFace \footnote{https://huggingface.co/datasets}.
These datasets can generally be divided into several categories such as web, academic, code, math, and other fields, with the majority of this data also being utilized in other representative LLMs. Table \ref{tab_data} details the statistics and sampling proportions of the data from different sources.
What we need to note is that we refer to the data not collected from web crawling and math as 'En-others' in order to uniformly adjust the weights. The specific datasets included are also provided in the Table \ref{tab_data}.
To ensure the quality of pretraining data we employ document-level Minhash deduplication and LSH algorithms to deduplicate all the aforementioned data.

Existing truly open-source language models still have room for improvement in Chinese contexts. Therefore, we have collected a large amount of Chinese open-source data online, including Telechat \citep{wang2024telechat}, Wudao \citep{yuan2021wudaocorpora}, Wanjuan \cite{he2023wanjuan}, Skypile \citep{wei2023skywork}, non-web data in MAP-CC \citep{du2024chinese} as well as the Chinese portions from multilingual corpora like Oasis, Oscar \citep{suarez2020monolingual} and CulturalX \citep{nguyen2023culturax}.
To further boost the quality of the Chinese data, we filtered all the Chinese data based on rules and language model perplexity scores. After filtering out low-quality data, we also performed deduplication on the Chinese data in the same way as for the English data.

Additionally, following \citep{bellagente2024stable,yuan2022restructured}, we have transformed multiple raw datasets into structured formats tailored for downstream tasks like question-answering, summarization, and sentiment analysis. Additionally, we have incorporated instructional data in \citep{longpre2023data}, and the aggregated dataset is referred to as Instruct in Table \ref{tab_data}. The list of sources used in Instruct are presented in Table \ref{tab_instruct}.

\subsection{Training}
We train our model from scratch with a context length of 4096 and adopt several efficient implementations to improve training speed. First, we use FlashAttention-2 \citep{dao2023flashattention} during training to increase device throughput.
Training is carried out using BFloat16 mixed precision while keeping all-reduce operations in FP32.
Our model is trained using the standard Adam optimizer with the following hyper-parameters: $\beta_1$ at 0.9 and $\beta_2$ at 0.95.
Additionally, we use a cosine learning rate schedule with a maximum learning rate of $3.0 \times 10-4$ and a minimum learning rate of $3.0 \times 10-5$.
The number of warmup steps is set to 4,000. 
Drawing from the detailed experimental analysis in \citet{bi2024deepseek}, we set the batch size to 5,242,880 tokens.
We assign weight decay as 0.1 and use a gradient clipping threshold of 1.0 to regulate the gradient value. 

\citet{muennighoff2024scaling} suggests that training for multiple epochs on repeated data results in negligible differences in loss compared to using unique data. 
We train on 2 trillion (2T) tokens for 2 epochs in this work.
To further enhance performance, after training on 4T tokens, we cool down the learning rate from 3e-5 to 0 which is similar to \citet{zhai2022scaling,bellagente2024stable}. At the same time, we reset the attention mask at the position of the end-of-sequence (EOS), as discussed in Section 2.1.

\section{Results}

In this section, we present the experimental results for ChuXin 1.6B on a wide range of benchmarks in both English and Chinese.
We compare ChuXin with other publicly available models on the same scale including Gemma \citep{team2024gemma}, StableLM 2\citep{bellagente2024stable}, H2O-Danube\citep{singer2024h2o}, Qwen1.5\citep{bai2023qwen}, TinyLLama\citep{zhang2024tinyllama}, Chinese Tiny LLM \citep{du2024chinese} and OLMo\citep{groeneveld2024olmo}.
We use the Language Model Evaluation Harness \citep{gao2021framework} to run evaluations and use the same evaluation metric with \citet{biderman2023pythia} for a fair comparison.

\paragraph{Common Sense Reasoning and Reading Comprehension.}
We evaluate our models on standard common sense reasoning and reading comprehension benchmarks, namely ARC easy and challenge \citep{clark2018think}, BooQ\citep{clark2019boolq}, Copa, Hellaswag\citep{zellers2019hellaswag}, OpenbookQA\citep{mihaylov2018can}, PIQA \citep{bisk2020piqa}, SciQ \citep{welbl2017crowdsourcing} and WinoGrande \citep{sakaguchi2021winogrande} in the zero-shot setting.

\begin{table*}[!ht]
\centering
\small
\begin{spacing}{1.2}
\setlength{\tabcolsep}{1.3mm}{
\begin{tabular}{lccccccccccc}
\toprule
{\bf Model} & {Size} & {\bf ARC-c} & {\bf ARC-e} & {\bf Boolq} & {\bf Copa} & {\bf Hella} & {\bf Open} & {\bf Piqa} & {\bf Sciq} & {\bf Wino} & {\bf Avg} \\
\midrule
Gemma & 2B & 48.98 & 78.45 & 69.51 & 84 & 71.73 & 39.8 & 78.02 & 94.3 & 65.51 & 70.03\\
H2O-Danube \dag & 1.8B & 35.84 & 62.29 & 65.81 & - & 68.20 & 37.6 & 76.93 & - & 61.96 & -\\
Qwen1.5 & 1.8B & 37.03 & 67.51 & 66.64 & 78 & 61.60 & 34.40 & 73.99 & 93 & 61.56	& 63.74 \\
StableLM 2 & 1.6B & 43.52 & 69.44 & 75.5 & 84 & 70.30 & 39.60 & 76.82 & 96.1 & 64.17 & 68.82 \\
\midrule
OpenLlama \dag & 3B & 34.00 & 69.00 & 68.00 & - & 49.00 & 40.00 & 75.00 & - & 62.00 & -\\
CT-LLM & 2B & 34.81 & 65.49	& 62.45	& 74 & 54.77 & 33.40 & 71.38 & 90.60 & 57.85 & 60.53 \\
TinyLLama & 1.1B & 34.81 & 67.47 & 63.15 & 74 & 60.00 & 34.60 & 73.12 & 88.8 & 58.88 &61.64 \\
OLMo & 1B & 34.22 & 67.55 & 61.40 & 82 & 63.96 & 36.40 & 75.10 & 86.70 & 60.30 & 63.07 \\
\midrule
ChuXin wo CD & 1.6B & 41.30 & 71.04 & 69.02 & 81 & 66.05 & 34.00 & 76.50 & 93.90 & 62.67 & 66.16 \\
ChuXin  & 1.6B & 39.68 & 71.38 & 71.25 & 83 & 66.09 & 35.00 & 77.09 & 95.00 & 63.54 & 66.89 \\
\bottomrule 
\end{tabular}}
\end{spacing}
\caption{
\label{tab1}
{\bf Performance on Common Sense Reasoning and Reading Comprehension tasks.}
Models with {\dag} denote that we directly report the scores from the corresponding paper, and others are from our implementation. Note that CD denotes cool down phase.
}
\end{table*}
\begin{table*}[!ht]
\centering
\small
\begin{spacing}{1.2}
\setlength{\tabcolsep}{1.3mm}{
\begin{tabular}{lccccccccc}
\toprule
{\bf Model} & {Size} & {\bf ARC} & {\bf Hella} & {\bf MMLU} & {\bf TQA} & {\bf Wino} & {\bf GSM} & {\bf Avg} & {\bf Avg wo GSM} \\
\midrule
Gemma & 2B & 48.98 & 71.73 & 42.47 & 33 & 65.51 & 10.08 & 45.3 & 52.34 \\
H2O-Danube & 1.8B & 39.68 & 69.75 & 25.97 & 33.63 & 64.17 & 2.05 & 39.21 & 46.64 \\
Qwen1.5 \dag & 1.8B & 37.88 & 61.42 & 46.71 & 39.43 & 60.3 & 33.59 & 46.55 & 49.15 \\
StableLM 2 & 1.6B & 43.52 & 70.30 & 39.8 & 36.61 & 64.17 & 17.29 & 45.28 & 50.88 \\
\midrule
OpenLlama \dag & 3B & 39.90 & 71.60 & 27.10 & 34.80 & 67.00 & 0.90 & 40.30 & 48.08 \\
CT-LLM & 3B & 34.81 & 54.77 & 37.81 & 39.81 & 57.85 & 7.35 & 38.73 & 45.01 \\
TinyLlama & 1.1B & 33.87 & 60.31 & 26.04 & 37.32 & 59.51 & 1.44 & 36.42 & 43.41 \\
Olmo & 1B & 34.22 & 63.96 & 26.03 & 32.9 & 60.3 & 1.44 & 36.48 & 43.48 \\
\midrule
ChuXin wo CD & 1.6B & 41.30 & 66.05 & 35.44 & 35.53 & 62.67 & 9.86 & 41.81 & 48.2 \\
ChuXin & 1.6B &  39.68 & 66.09 & 41.07 & 37.65 & 63.54 & 12.66 & 43.45 & 49.61 \\
\bottomrule 
\end{tabular}}
\end{spacing}
\caption{
\label{tab2}
{\bf Performance on Open LLM Leaderboard.}
Models with {\dag} denote that we directly report the scores from the Open LLM Leaderboard, and others are from our implementation. For every model listed in the table, we present the scores for each separate benchmark, along with the average score and the average score excluding the GSM8k benchmark.
}
\end{table*}

Table \ref{tab1} shows the results of common sense reasoning and reading comprehension benchmarks.
On these benchmarks, ChuXin achieved competitive results with existing open-source models of the same size.
Compared with existing models and fully open-sourced language models like tiny-llama \citep{zhang2024tinyllama} and olmo \citep{groeneveld2024olmo}, ChuXin has achieved significant performance improvement. The enhancement in performance primarily comes from more training data and a small increase in the scale of model parameters.
Compared to models with closer parameter scales like Gemma \citep{team2024gemma}, StableLM \citep{bellagente2024stable}, and Qwen1.5 \citep{bai2023qwen}, ChuXin still has a certain gap on some benchmarks. However, since the training data for these models is not completely public, the performance gap could potentially stem from the data.
From table \ref{tab1}, we can also see that during the cooldown phase, the average performance of ChuXin improved by one points.

\paragraph{Open LLM Leaderboard.} Open LLM Leaderboard \footnote{https://huggingface.co/spaces/HuggingFaceH4/open\_llm\_leaderboard} evaluates pretrain models on 6 key benchmarks which are designed to examine various types of reasoning and general knowledge spanning numerous domains, in both zero-shot and few-shot evaluation settings. They are ARC challenge (25-shot) \citep{clark2018think} HellaSwag (10-shot) \citep{zellers2019hellaswag}, MMLU (5-shot) \citep{hendrycks2020measuring}, TruthfulQA (0-shot) \citep{lin2021truthfulqa}, Winogrande (5-shot) \citep{sakaguchi2021winogrande}, GSM8k (5-shot) \citep{cobbe2021training}. 
At evaluation time, we use the examples provided by the benchmark, and the results of our models on the benchmark are reported in Table \ref{tab2}. 

We observe that ChuXin performed comparably with Gemma, StableLM, and Qwen1.5, and its performance is even closer to these models when excluding the GSM8K test set.
One possible reason for this behavior could be the specialized data utilized in training the Qwen1.5 and Stable LM 2 models, which enhanced performance on certain benchmarks. For instance, the gsm8k-ScRel dataset was employed by Qwen1.5 to improve mathematical reasoning capabilities \citep{bai2023qwen}.

\begin{table*}[!ht]
\centering
\small
\begin{spacing}{1.2}
\setlength{\tabcolsep}{1.3mm}{
\begin{tabular}{lcccc}
\toprule
{\bf Model} & {Size} & {\bf C-Eval} & {\bf CMMLU} & {\bf Humaneval} \\
\midrule
Stable LM 2 & 1.6B & 29.27 & 30.10 & 7.32  \\
Gemma & 2B & 31 & 31.06 & 9.51  \\
Qwen1.5 & 1.8B & 59.38 & 57.08 & 23.17  \\
CT-LLM & 2B & 36.78 & 36.40 & 9.15 \\
\midrule
ChuXin  & 1.6B & 39.31 & 37.11 & 9.76 \\
\bottomrule 
\end{tabular}}
\end{spacing}
\caption{
\label{tab3}
{\bf Performance on CMMLU, C-Eval and HumanEval.} We report pass@1 scores on HumanEval.
}
\end{table*}

\paragraph{Chinese Evaluation.}
We use two Chinese understanding and reasoning test sets, the CMMLU \citep{li2023cmmlu} and the C-Eval \citep{huang2024c} to evaluate the model's capability in Chinese tasks. At the same time, we assess the model's code generation ability using the HumanEval \citep{chen2021evaluating}.
Table \ref{tab3} displays the performance of all models on the two Chinese test sets. From the table, we can see that among models around the 2B scale, ChuXin's Chinese performance ranks second only after Qwen1.5.
Since Stable LM and Gemma use only a small amount of Chinese data during training (mixed in with multilingual data), their performance on Chinese tasks lags behind models that have a higher proportion of Chinese pre-training data.
In terms of code completion ability, ChuXin achieves comparable performance with models of the same size, but there is still a gap compared to Qwen1.5.

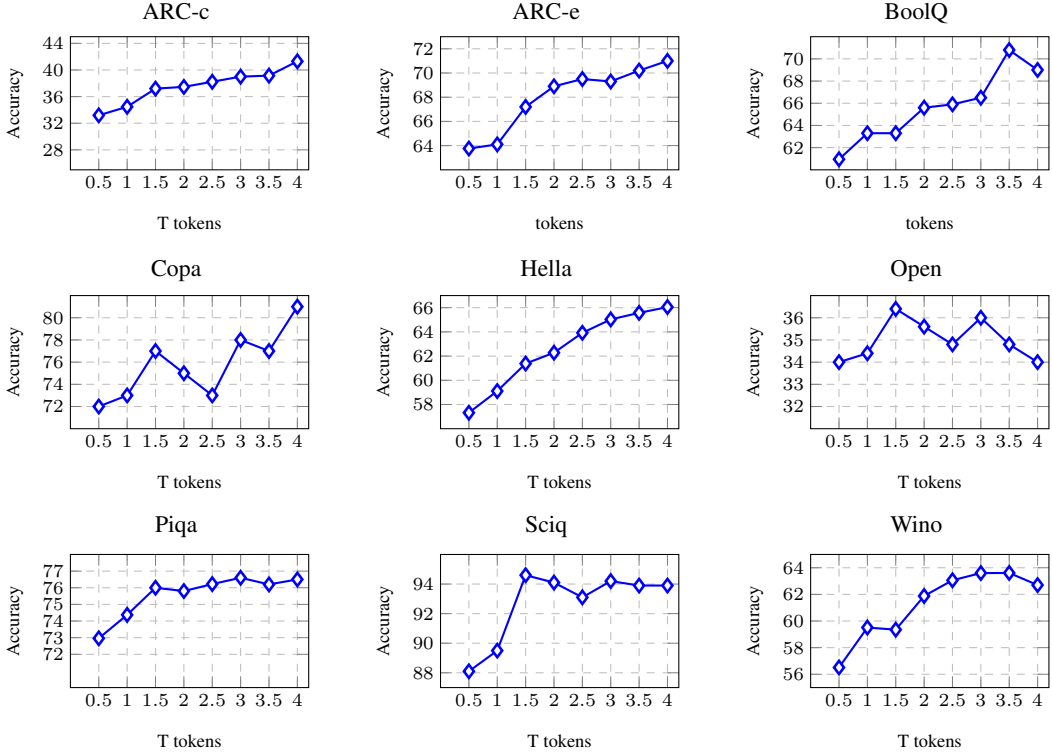
\begin{figure*}[!t] 
  \centering
  \begin{tikzpicture}[]
  \node[inner sep=0pt,font=\footnotesize] (d1) at (4em,6em){ARC-c};
  \node[inner sep=0pt,font=\footnotesize,anchor=north] (d2) at ([xshift=14em]d1.north){ARC-e};
  \node[inner sep=0pt,font=\footnotesize,anchor=north] (d3) at ([xshift=14em]d2.north){BoolQ};
  \node[inner sep=0pt,font=\footnotesize,anchor=north] (d4) at ([yshift=-9.7em]d1.north){Copa};
  \node[inner sep=0pt,font=\footnotesize,anchor=north] (d5) at ([yshift=-9.7em]d2.north){Hella};
  \node[inner sep=0pt,font=\footnotesize,anchor=north] (d6) at ([yshift=-9.7em]d3.north){Open};
  \node[inner sep=0pt,font=\footnotesize,anchor=north] (d7) at ([yshift=-9.7em]d4.north){Piqa};
  \node[inner sep=0pt,font=\footnotesize,anchor=north] (d8) at ([yshift=-9.7em]d5.north){Sciq};
  \node[inner sep=0pt,font=\footnotesize,anchor=north] (d9) at ([yshift=-9.7em]d6.north){Wino};
    \scriptsize{
      \begin{axis}[
	 at={(0,0)},
      ymajorgrids,
      xmajorgrids,
      grid style=dashed,
      width=.34\textwidth,
      height=.24\textwidth,
      legend style={at={(0.38,0.13)}, anchor=south west},
      xlabel={\scriptsize{T tokens}},
      ylabel={\scriptsize{Accuracy}},
      ylabel style={yshift=-2em},xlabel style={yshift=0.0em},
      ymin=25,ymax=45, ytick={28,32,36,40,44},
      xmin=0,xmax=4.2,xtick={0.5,1,1.5,2,2.5,3,3.5,4},
      ]

      \addplot[blue,mark=diamond*,mark size=2.5pt,thick,mark options={fill=white,draw=blue,line width=1.0pt}] coordinates {(0.5,33.19) (1,34.47) (1.5,37.2) (2,37.46) (2.5,38.23) (3,39.00) (3.5,39.15) (4,41.3)
      };
      \end{axis}
     }

    \scriptsize{
      \begin{axis}[
	 at={(20em,0)},
      ymajorgrids,
      xmajorgrids,
      grid style=dashed,
      width=.34\textwidth,
      height=.24\textwidth,
      legend style={at={(0.38,0.13)}, anchor=south west},
      xlabel={\scriptsize{tokens}},
      ylabel={\scriptsize{Accuracy}},
      ylabel style={yshift=-2em},xlabel style={yshift=0.0em},
      ymin=62,ymax=73, ytick={64,66,68,70,72},
      xmin=0,xmax=4.2,xtick={0.5,1,1.5,2,2.5,3,3.5,4},
      ]

      \addplot[blue,mark=diamond*,mark size=2.5pt,thick,mark options={fill=white,draw=blue,line width=1.0pt}] coordinates {(0.5,63.76) (1,64.1) (1.5,67.2) (2,68.9) (2.5,69.5) (3,69.3) (3.5,70.2) (4,71.0)
      };
      \end{axis}
     }

     \scriptsize{
      \begin{axis}[
	 at={(40em,0)},
      ymajorgrids,
      xmajorgrids,
      grid style=dashed,
      width=.34\textwidth,
      height=.24\textwidth,
      legend style={at={(0.38,0.13)}, anchor=south west},
      xlabel={\scriptsize{tokens}},
      ylabel={\scriptsize{Accuracy}},
      ylabel style={yshift=-2em},xlabel style={yshift=0.0em},
      ymin=60,ymax=72, ytick={62,64,66,68,70},
      xmin=0,xmax=4.2,xtick={0.5,1,1.5,2,2.5,3,3.5,4},
      ]

      \addplot[blue,mark=diamond*,mark size=2.5pt,thick,mark options={fill=white,draw=blue,line width=1.0pt}] coordinates {(0.5,60.95) (1,63.3) (1.5,63.3) (2,65.6) (2.5,65.9) (3,66.5) (3.5,70.8) (4,69.0)
      };
      \end{axis}
     }
     \scriptsize{
      \begin{axis}[
	 at={(0,-14em)},
      ymajorgrids,
      xmajorgrids,
      grid style=dashed,
      width=.34\textwidth,
      height=.24\textwidth,
      legend style={at={(0.38,0.13)}, anchor=south west},
      xlabel={\scriptsize{T tokens}},
      ylabel={\scriptsize{Accuracy}},
      ylabel style={yshift=-2em},xlabel style={yshift=0.0em},
      ymin=70,ymax=82, ytick={72,74,76,78,80},
      xmin=0,xmax=4.2,xtick={0.5,1,1.5,2,2.5,3,3.5,4},
      ]

      \addplot[blue,mark=diamond*,mark size=2.5pt,thick,mark options={fill=white,draw=blue,line width=1.0pt}] coordinates {(0.5,72) (1,73) (1.5,77) (2,75) (2.5,73) (3,78) (3.5,77) (4,81)
      };
      \end{axis}
     }
          \scriptsize{
      \begin{axis}[
	 at={(20em,-14em)},
      ymajorgrids,
      xmajorgrids,
      grid style=dashed,
      width=.34\textwidth,
      height=.24\textwidth,
      legend style={at={(0.38,0.13)}, anchor=south west},
      xlabel={\scriptsize{T tokens}},
      ylabel={\scriptsize{Accuracy}},
      ylabel style={yshift=-2em},xlabel style={yshift=0.0em},
      ymin=56,ymax=67, ytick={58,60,62,64,66},
      xmin=0,xmax=4.2,xtick={0.5,1,1.5,2,2.5,3,3.5,4},
      ]

      \addplot[blue,mark=diamond*,mark size=2.5pt,thick,mark options={fill=white,draw=blue,line width=1.0pt}] coordinates {(0.5,57.32) (1,59.11) (1.5,61.38) (2,62.28) (2.5,63.92) (3,65.03) (3.5,65.57) (4,66.05)
      };
      \end{axis}
     }
          \scriptsize{
      \begin{axis}[
	 at={(40em,-14em)},
      ymajorgrids,
      xmajorgrids,
      grid style=dashed,
      width=.34\textwidth,
      height=.24\textwidth,
      legend style={at={(0.38,0.13)}, anchor=south west},
      xlabel={\scriptsize{T tokens}},
      ylabel={\scriptsize{Accuracy}},
      ylabel style={yshift=-2em},xlabel style={yshift=0.0em},
      ymin=31,ymax=37, ytick={32,33,34,35,36},
      xmin=0,xmax=4.2,xtick={0.5,1,1.5,2,2.5,3,3.5,4},
      ]

      \addplot[blue,mark=diamond*,mark size=2.5pt,thick,mark options={fill=white,draw=blue,line width=1.0pt}] coordinates {(0.5,34) (1,34.4) (1.5,36.4) (2,35.6) (2.5,34.8) (3,36.0) (3.5,34.8) (4,34)
      };
      \end{axis}
     }
     \scriptsize{
      \begin{axis}[
	 at={(0,-28em)},
      ymajorgrids,
      xmajorgrids,
      grid style=dashed,
      width=.34\textwidth,
      height=.24\textwidth,
      legend style={at={(0.38,0.13)}, anchor=south west},
      xlabel={\scriptsize{T tokens}},
      ylabel={\scriptsize{Accuracy}},
      ylabel style={yshift=-2em},xlabel style={yshift=0.0em},
      ymin=70,ymax=78, ytick={72,73,74,75,76,77},
      xmin=0,xmax=4.2,xtick={0.5,1,1.5,2,2.5,3,3.5,4},
      ]

      \addplot[blue,mark=diamond*,mark size=2.5pt,thick,mark options={fill=white,draw=blue,line width=1.0pt}] coordinates {(0.5,72.96) (1,74.37) (1.5,76.0) (2,75.8) (2.5,76.22) (3,76.6) (3.5,76.2) (4,76.5)
      };
      \end{axis}
     }
          \scriptsize{
      \begin{axis}[
	 at={(20em,-28em)},
      ymajorgrids,
      xmajorgrids,
      grid style=dashed,
      width=.34\textwidth,
      height=.24\textwidth,
      legend style={at={(0.38,0.13)}, anchor=south west},
      xlabel={\scriptsize{T tokens}},
      ylabel={\scriptsize{Accuracy}},
      ylabel style={yshift=-2em},xlabel style={yshift=0.0em},
      ymin=87,ymax=96, ytick={88,90,92,94},
      xmin=0,xmax=4.2,xtick={0.5,1,1.5,2,2.5,3,3.5,4},
      ]

      \addplot[blue,mark=diamond*,mark size=2.5pt,thick,mark options={fill=white,draw=blue,line width=1.0pt}] coordinates {(0.5,88.1) (1,89.5) (1.5,94.6) (2,94.1) (2.5,93.1) (3,94.2) (3.5,93.9) (4,93.9)
      };
      \end{axis}
     }
          \scriptsize{
      \begin{axis}[
	 at={(40em,-28em)},
      ymajorgrids,
      xmajorgrids,
      grid style=dashed,
      width=.34\textwidth,
      height=.24\textwidth,
      legend style={at={(0.38,0.13)}, anchor=south west},
      xlabel={\scriptsize{T tokens}},
      ylabel={\scriptsize{Accuracy}},
      ylabel style={yshift=-2em},xlabel style={yshift=0.0em},
      ymin=55,ymax=65, ytick={56,58,60,62,64},
      xmin=0,xmax=4.2,xtick={0.5,1,1.5,2,2.5,3,3.5,4},
      ]

      \addplot[blue,mark=diamond*,mark size=2.5pt,thick,mark options={fill=white,draw=blue,line width=1.0pt}] coordinates {(0.5,56.51) (1,59.51) (1.5,59.35) (2,61.88) (2.5,63.06) (3,63.6) (3.5,63.61) (4,62.7)
      };
      \end{axis}
     }
  \end{tikzpicture}
  \caption{Performance on commonsense reasoning benchmarks during pre-training}\label{fig1}
\end{figure*}

\paragraph{Performance during Training}
We monitored the performance of ChuXin on commonsense reasoning benchmarks throughout its pre-training phase, as depicted in Fig \ref{fig1}. From the figure, we can see that the performance of ChuXin on most tasks (except for OpenbookQA) improves with the increase in the number of training tokens, which also corroborates the conclusion in the \citet{muennighoff2024scaling}.

\begin{figure}[!ht] 
\centering 
\includegraphics[scale=0.5]{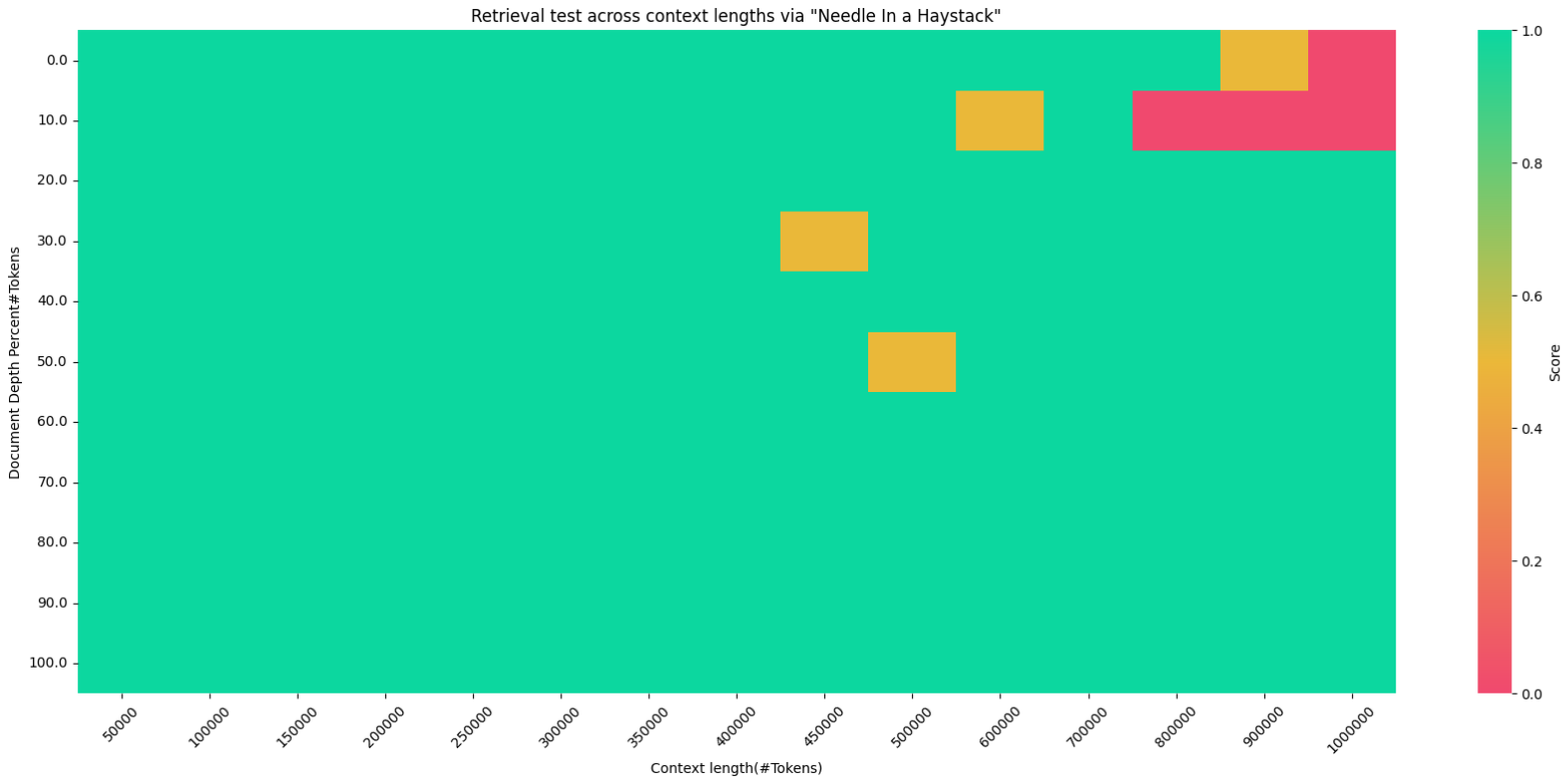} 
\caption{Retrieval test on ChuXin-1M across context lengths via "Needle In a Haystack"} 
\label{figure2} 
\end{figure}

\begin{table*}[!ht]
\centering
\small
\begin{spacing}{1.2}
\setlength{\tabcolsep}{1.3mm}{
\begin{tabular}{lccccccccccc}
\toprule
{\bf Model}  & {\bf ARC-c} & {\bf ARC-e} & {\bf Boolq} & {\bf Copa} & {\bf Hella} & {\bf Open} & {\bf Piqa} & {\bf Sciq} & {\bf Wino} & {\bf Avg} \\
ChuXin-base   & 39.68 & 71.38 & 71.25 & 83 & 66.09 & 35.00 & 77.09 & 95.00 & 63.54 & 66.89 \\
ChuXin-32k   &  39.16 & 70.66 & 67.71 & 81 & 65.69   & 35.8   & 76.88  & 94.2  & 62.51 & 65.96 \\
ChuXin-64k  & 38.48 &  70.24  & 67.52  & 82  & 65.6   & 35.2 & 76.61 & 94.3 & 63.3  & 65.92 \\
ChuXin-128k  & 39.08 & 69.4  & 67.71  &  80  & 65.74 & 35.4 & 76.39 & 94.1 & 63.3 & 65.68  \\
ChuXin-256k  & 40.19 & 70.75  & 69.3  & 78  & 65.85  & 35.8 & 76.88 & 93.5  & 63.85 & 66.01 \\
ChuXin-512k  & 40.61 & 71.21 & 67.77  & 78 & 64.82 & 34.8 & 76.88 & 93.6  & 61.88 & 65.51 \\
ChuXin-1M  & 41.13 & 72.26 & 62.08 & 75 & 64.59 & 34.8 & 76.71 & 93.33 & 62.43 & 64.7 \\
\bottomrule 
\end{tabular}}
\end{spacing}
\caption{
\label{tab4}
{\bf Performance of the models trained under different contextual windows on Common Sense Reasoning and Reading Comprehension tasks}
}
\end{table*}

\begin{table*}[!ht]
\centering
\small
\begin{spacing}{1.2}
\setlength{\tabcolsep}{1.3mm}{
\begin{tabular}{lccccccccc}
\toprule
{\bf Model} & {\bf ARC} & {\bf Hella} & {\bf MMLU} & {\bf TQA} & {\bf Wino} & {\bf GSM} & {\bf Avg} & {\bf Avg wo GSM} \\
ChuXin-base & 39.68 & 66.09 & 41.07 & 37.65 & 63.54 & 12.66 & 43.45 & 49.61 \\
ChuXin-32k  &  39.16 & 65.69 & 38.63 & 35.66 & 62.51 & 11.6 & 42.21 & 48.33 \\
ChuXin-64k &  38.48 & 65.6 & 38.43 & 35.07 & 63.3 & 11.9 & 42.13 & 48.18 \\
ChuXin-128k  &  39.08 & 65.74 & 37.65 & 34.89  & 63.3  & 11.07 & 41.96 & 48.13 \\
ChuXin-256k  &  40.19 & 65.85 & 37.16 & 35.2 & 63.85 & 10.16 & 42.07 & 48.45 \\
ChuXin-512k  &  40.61 & 64.82 & 36.66 & 33.66 & 61.88 & 8.11 & 40.96 & 47.53 \\
ChuXin-1M  &  41.13 & 64.59 & 35.76 & 34.67 & 62.43 & 6.82 & 40.9 & 47.72 \\
\bottomrule 
\end{tabular}}
\end{spacing}
\caption{
\label{tab5}
{\bf Performance of the models trained under different contextual windows on Open LLM Leaderboard}
}
\end{table*}

\section{Long Context Extension}
Following the initial training phase of the ChuXin, we implement the Adjusted Base Frequency (ABF) technique and adopted a curriculum learning approach to progressively extend the context window from 4,000 tokens to 1 million tokens.
Inspired by \citet{fu2024data}, we upsampled the data of long sequences and incorporated a small portion of synthesized long-sequence data for training. Additionally, we designed a new parallel training framework based on distributed attention to efficiently support the training of long sequences. Table \ref{tab4} and  \ref{tab5} show the performance of the models trained under different contextual windows on downstream tasks, demonstrating that expanding the context size does not severely compromise short context task performance. And as shown in Figure \ref{figure2}, the results on the "Needle In A Haystack"(NIAH) tests indicate that ChuXin-1M performs well across all context window lengths up to 1M.

\section{Conclusion and Future Work}
In this work, we introduce ChuXin, a fully open-source model along with all the details needed to train it, including the training data, model structure, and hyperparameters. 
Additionally, we have successfully expanded the model's context capacity to accommodate up to 1 million tokens by applying a method of efficient, ongoing pre-training. This milestone marks the beginning of our commitment to open-source initiatives. Looking ahead, our roadmap includes the release of larger and even more capable models that feature instruction tuning and integration of diverse modalities. We will also chronicle the challenges we faced during the development of ChuXin and the strategies we employed to overcome them, contributing valuable insights to the open-source community and fostering further advancements in the domain of language modeling.

\bibliography{iclr2023_conference}
\bibliographystyle{iclr2023_conference}

\appendix
\section{Appendix}
\begin{table*}[!ht]
\centering
\small
\begin{spacing}{1.1}
\setlength{\tabcolsep}{4mm}{
\begin{tabular}{ll}
\toprule
{\bf Dataset} & {\bf Prefix} \\
\midrule
Banking77 & banking77 \\
BigPatent & big\_patent \\
CLOTH & AndyChiang/cloth \\
CommonGen & common\_gen \\
FigQA & nightingal3/fig-qa \\
Flan & 2021 DataProvenanceInitiative/flan2021\_submix\_original \\
Flan & NIv2 DataProvenanceInitiative/niv2\_submix\_original \\
HelpSteer & nvidia/HelpSteer \\
IMDB & Review ajaykarthick/imdb-movie-reviews \\
Joke & Explanation dim/joke\_explaination \\
SciTLDR & allenai/scitldr \\
Self-Instruct Starcoder & codeparrot/self-instruct-starcoder \\
SQL Create Context & b-mc2/sql-create-context \\
StepGame & tasksource/stepgame \\
TRACIE & tasksource/tracie \\ 
WordNet & jon-tow/open-english-wordnet-synset-2023 \\
Yahoo Answers Topics & yahoo\_answers\_topics \\
\bottomrule 
\end{tabular}}
\end{spacing}
\caption{
\label{tab_instruct}
 To locate the base datasets used for restructured and instruction-based pre-training, visit https://huggingface.co/datasets/ and append the given prefix to the URL. }
\end{table*}

\end{document}

%% file: math_commands.tex
\usepackage{amsmath,amsfonts,bm}

\def\eqref#1{equation~\ref{#1}}

\def\1{\bm{1}}

\DeclareMathAlphabet{\mathsfit}{\encodingdefault}{\sfdefault}{m}{sl}
\SetMathAlphabet{\mathsfit}{bold}{\encodingdefault}{\sfdefault}{bx}{n}

